\documentclass[11pt]{article}

\usepackage[final]{acl} 

\usepackage{times}
\usepackage{latexsym}

\usepackage[T1]{fontenc}

\usepackage[utf8]{inputenc}
\usepackage{microtype}
\usepackage{inconsolata}
\usepackage{graphicx}

\usepackage{fontawesome5}
\usepackage{amsmath}
\usepackage{amssymb}
\usepackage[ruled, lined, noend]{algorithm2e}
\usepackage{multirow}

\usepackage[nonfrench,finemath]{kotex} 
\usepackage{dhucs-nanumfont}           

\usepackage{xcolor}
\usepackage{pifont}

\usepackage{placeins}

\definecolor{green1}{RGB}{0,148,0}
\definecolor{red1}{RGB}{180,0,0}

\newcommand{\cmark}{\ding{51}} 
\newcommand{\xmark}{\ding{55}} 

\newcommand{\green}[1]{\textcolor{green1}{#1}}
\newcommand{\red}[1]{\textcolor{red1}{#1}}

\title{\faBomb\ REZE: Representation Regularization for Domain-adaptive Text Embedding Pre-finetuning}

\author{
Seungmin Lee\thanks{~~Equal contribution. Author order was decided by rock-paper-scissors.}\thanks{~~Work completed as an intern at PwC Korea GenAI Team.} \quad 
  Jeonghwan Lee\footnotemark[1]\thanks{~~Corresponding author.} \quad
  Hyunkuk Lim\ \quad 
  Sejoon Kim \quad 
  Mingi Sung\\
  PwC Korea GenAI Team, Seoul, South Korea \\
  \texttt{elplaguister@yonsei.ac.kr} \\
  \texttt{\{jeonghwan.lee, hyunkuk.lim, sejoon.s.kim, mingi.sung\}@pwc.com}
}

\begin{document}
\maketitle
\begin{abstract}

Recent text embedding models are often adapted to specialized domains via contrastive pre-finetuning (PFT) on a naive collection of scattered, heterogeneous tasks. However, this approach often introduces task-induced bias alongside domain knowledge, leading to uncontrolled representation shifts that distort the pretrained embedding geometry and cause substantial performance degradation.
To address this issue, we propose \textbf{REZE}, a \underline{\textbf{re}}presentation regulari\underline{\textbf{z}}ation framework that explicitly controls representation shift during \underline{\textbf{e}}mbedding pre-finetuning. REZE operates on the relations of anchor-positive pairs and decomposes them in an eigenspace. It then measures task-wise dispersion along each eigencomponent to identify task-variant directions and applies adaptive soft-shrinkage to suppress task-induced noise while preserving task-invariant semantic structure, without inference-time overhead. Experiments across multiple embedding backbones and specialized benchmarks show that REZE outperforms standard pre-finetuning and isotropy-oriented post-hoc regularization in most settings, remaining stable where existing PFT variants collapse. Embedding space analyses further confirm that REZE induces controlled shifts aligned with the original embedding manifold, underscoring representation shift control as a key principle for robust embedding pre-finetuning under heterogeneous supervision.

\end{abstract}

\section{Introduction}

\begin{figure}[t]
  \includegraphics[width=\columnwidth]{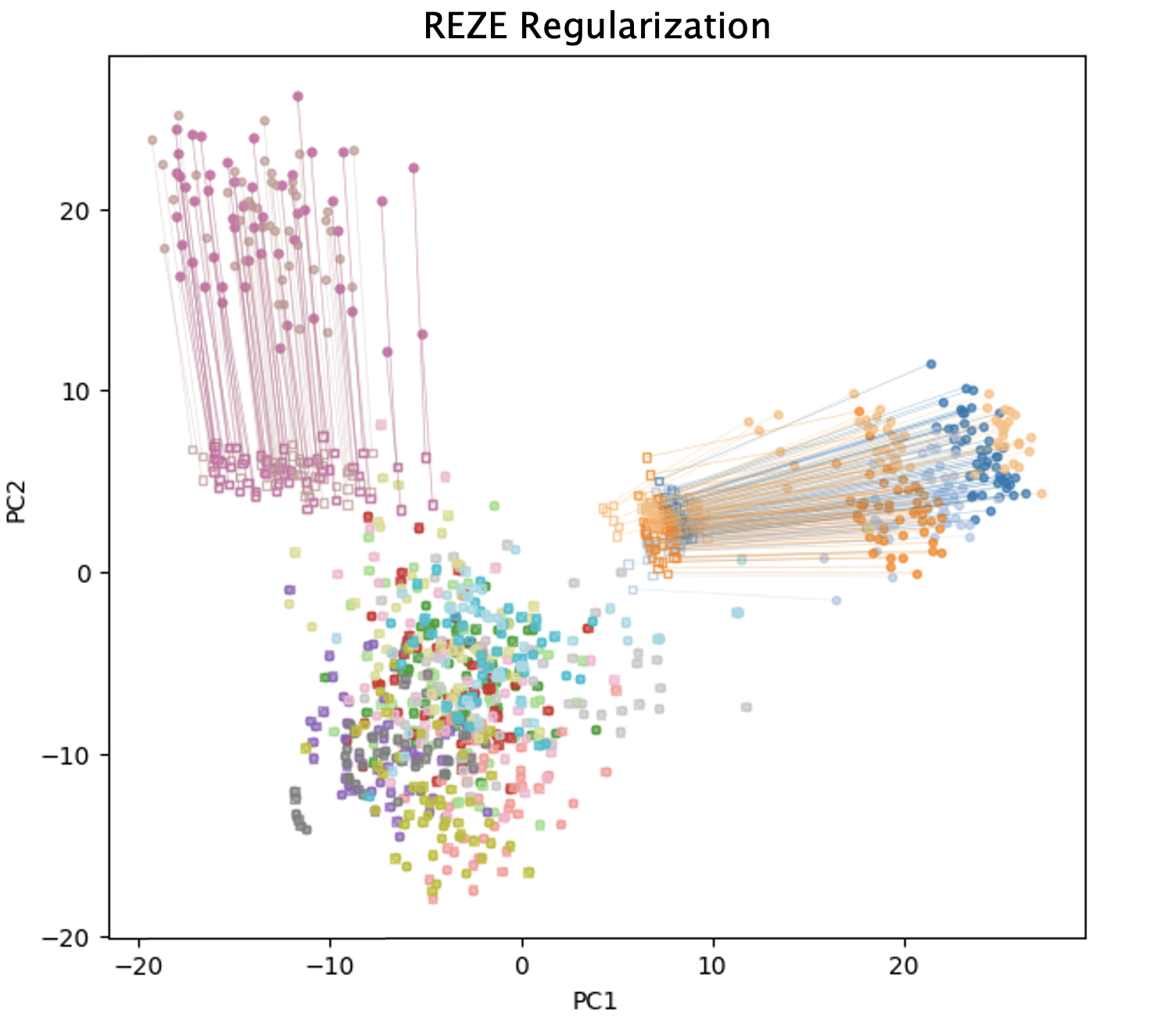}
\caption{Visualization of REZE’s effect on representation geometry.
REZE suppresses task-variant deviations by softly shrinking source-separating components toward a global reference, thereby reducing dataset-specific artifacts that lead to task conflicts.
Darker points denote the original embeddings, while lighter points of the same color indicate the corresponding embeddings after REZE regularization.
Each pair of points connected by a line corresponds to the same text instance, illustrating how REZE adjusts representations in the eigenspace.}
  \label{fig:experiments}
\end{figure}

\subsection{Background and Motivation}
Modern general-purpose embedding models have achieved remarkable success across diverse downstream tasks by leveraging multi-stage training and weak supervision at scale \citep{wang2022text,li2023towards}. A critical component in this pipeline is pre-finetuning—an intermediate training stage that serves a dual purpose: adapting the original pre-training objective (e.g., masked language modeling) into one specialized for embedding tasks, and acting as a warm-up phase that prepares the model for downstream fine-tuning.

However, there is a growing demand for adapting general-purpose embedding models to specialized domains. Ideally, such adaptation would rely on large-scale domain-specific corpora, but in reality, these are often scarce or unavailable. Practitioners are thus compelled to collect and aggregate smaller, fragmented datasets scattered throughout the domain (e.g., classification, retrieval, NLI). While this heterogeneous collection offers a practical means to infuse domain knowledge, the fundamental dissimilarity among tasks introduces potential task conflicts and negative transfer, indicating that naive multi-task learning alone is insufficient for effective domain knowledge transfer.

A potential solution lies in representation regularization, which we define as the systematic removal of task-induced biases from learned representations, with the objective of preserving only the generalizable domain knowledge shared among diverse tasks. It is well-established that isotropic representations lead to improved performance across diverse downstream tasks \citep{balestriero2025lejepaprovablescalableselfsupervised}. Consequently, there have been efforts to transform anisotropic representations into isotropic ones through learning or post-processing methods, with "whitening" being a representative example \citep{su2021whitening}. Such techniques span a wide spectrum of implementations, ranging from post-processing methods to additional trainable layers, and further extending to approaches that require no auxiliary parameters during fine-tuning. To the best of our knowledge, the majority of existing methodologies have been validated only in single-task settings, and no prior work has systematically investigated configurations involving multiple heterogeneous datasets, such as those encountered in pre-finetuning scenarios for domain-specific embedding models.

\subsection{Proposed Approach}
We aim to constrain the growth of task-induced biases in the embedding space via component-wise reconstruction in the latent space. To this end, we decouple task-specific knowledge from domain-common knowledge across multiple datasets. Our objective is to preserve the generalizable domain knowledge while mitigating the task-specific biases embedded within individual datasets.

It should be noted that distinguishing between task noise and genuinely useful information within dataset bias is inherently difficult to determine a priori. Therefore, in our proposed auxiliary objective function, learning proceeds in a direction that reduces overall task bias, while the main loss function is designed to leverage the actually useful information.

To realize this objective, we propose REZE, a representation regularization framework designed to control representation shifts during embedding pre-finetuning. As illustrated in Figure \ref{fig:experiments}, REZE aims to filter out task-induced noise and retain domain-general knowledge, while preventing unnecessary distortion of the representation space.
Instead of operating on individual embeddings, our approach focuses on relation representations, which we define as the concatenation of anchor and positive embedding pairs. Our key insight is that while generalizable domain knowledge is shared across heterogeneous tasks, task-specific biases manifest as outliers in the geometric structure of these relations. REZE decomposes these representations into an eigenspace to disentangle these components. It then employs a robust statistic-based noise detection mechanism to identify task-variant directions and applies adaptive soft-shrinkage to selectively suppress them. This process guides the model to learn a more robust embedding space that is less susceptible to negative transfer from conflicting tasks.

\section{Related Work}
\subsection{Embedding Model and Pre-Finetuning}
Text embedding research has evolved from static lexical representations such as word2vec and GloVe \citep{mikolov2013efficient,pennington-etal-2014-glove}, which lack context sensitivity, to contextual Transformer-based encoders. Modern approaches largely adopt a bi-encoder paradigm where texts are encoded independently and compared via simple similarity functions. Key milestones include Sentence-BERT \citep{reimers2019sentence}, DPR \citep{karpukhin2020dense}, and SimCSE \citep{gao-etal-2021-simcse}, which collectively established contrastive training as the dominant framework for sentence embeddings.

Recent advancements in general-purpose embedding models have established pre-finetuning as a standard intermediate stage in the training pipeline. Studies such as \citealp{wang2022text} and \citealp{li2023towards} demonstrate the efficacy of this stage, which adapts the original pre-training objective (e.g., masked language modeling) into one specialized for embedding tasks. This phase acts as a crucial warm-up, preparing the model for effective downstream fine-tuning by exposing it to diverse weak supervision signals.

\subsection{Challenges and Performance Degradation in Multi-task Learning}
Multi-task Learning (MTL) aims to achieve superior performance compared to independent learning by leveraging shared information across multiple tasks. However, in practice, conflicts between tasks have been frequently reported to cause performance degradation, where individual task performance falls below that of Single-task Learning \citep{chai2023improving, zhang2024proactive}. These challenges include gradient conflict and gradient domination, where competing parameter updates across tasks hinder optimization, as well as negative transfer, where learning conflicting or unrelated tasks compromises overall generalization \citep{lee2024instruction, romero2025beyond}.

The primary focus of this work lies in eliminating task bias arising from heterogeneous tasks during the pre-finetuning stage, rather than in improving MTL itself. By proactively mitigating the inherent interference factors of heterogeneous tasks, we prevent the model from becoming overly biased toward any specific task. Through this approach, the core objective of our work is to enable the model to extract only the generalizable and shared knowledge that pervades the entire domain, rather than the individual characteristics of each task, and transfer this knowledge to downstream tasks.

\subsection{Representation regularization}
Sentence embeddings derived from pre-trained language models (PLMs) often exhibit anisotropy, where representations concentrate in a narrow region of the vector space. This can destabilize cosine-similarity comparisons and amplify dataset-specific biases or spurious correlations.
To mitigate this issue, BERT-flow \citep{li2020sentence} proposes learning an invertible normalizing flow to map embeddings into an isotropic Gaussian distribution.
WhiteningBERT \citep{huang2021whiteningbert} and \citealp{su2021whitening} show that a simple linear whitening transform \citep{friedman1987exploratory} can substantially improve isotropy without additional training.
 \citet{jung2023isotropic} further extends isotropization to dense retrieval by adopting both sequence-wise and token-wise transformations to improve retrieval representations.
From a contrastive-learning perspective, DCLR \citep{zhou2022debiased} improves uniformity by combining Gaussian-initialized noise-based negatives with instance weighting to down-weight false negatives.
More recently, Kernel-Whitening \citep{gao2022kernel} utilizes Nyström kernel approximation \citep{williams2000using} to address nonlinear spurious correlations, extending whitening-style regularization beyond linear transformations.

Besides, this work builds on the shared goal of stabilizing representation geometry, but targets task-variant components that emerge in heterogeneous multi-dataset pre-finetuning.

\section{REZE Regularization}
\label{sec:reze-regularization}

\subsection{Reference Eigenspace Construction}
\label{sec:offline-reze-matrix}

REZE aims to preserve task-invariant components of relation representations while softly suppressing task-variant components that encode task-specific biases in heterogeneous multi-corpus pre-finetuning.

Let $S$ be the number of source datasets used in pre-finetuning, indexed by $s\in\{1,\dots,S\}$.
Given anchor--positive pairs from each source dataset $s$, we form a relation representation by concatenating
$\mathbf{r}_{s,i}=[\mathbf{a}_{s,i};\mathbf{p}_{s,i}] \in \mathbb{R}^{D}$,
where $\mathbf{a}_{s,i}, \mathbf{p}_{s,i} \in \mathbb{R}^{d}$ denote the $d$-dimensional embedding vectors, and $D=2d$ is the combined dimensionality.
Then we aggregate all samples into a pooled set $\{\mathbf{x}_n\}_{n=1}^{N}$ with the source indicator $\mathrm{src}(n)\in\{1,\dots,S\}$,
where $N$ denotes the total number of pooled pre-finetuning samples across all sources.


\paragraph{Eigen Value Decomposition (EVD)}
We compute the global mean $\mathbf{u}=\frac{1}{N}\sum_{n=1}^{N}\mathbf{x}_n$ and center each sample $\tilde{\mathbf{x}}_n=\mathbf{x}_n-\mathbf{u}$.
Let $\tilde{\mathbf{X}}\in\mathbb{R}^{N\times D}$ be the centered matrix whose $n$-th row is $\tilde{\mathbf{x}}_n^\top$.
We construct the covariance $\mathbf{C}=\frac{1}{N}\tilde{\mathbf{X}}^\top\tilde{\mathbf{X}}$ and perform EVD
\begin{equation}
\mathbf{C}=\mathbf{W}\mathbf{\Lambda}\mathbf{W}^\top,
\qquad
\mathbf{\Lambda}=\mathrm{diag}(\lambda_1,\dots,\lambda_D),
\label{eq:reze-evd}
\end{equation}
which defines eigenspace representations $\mathbf{z}_n=\mathbf{W}^\top(\mathbf{x}_n-\mathbf{u})\in\mathbb{R}^{D}$.

\paragraph{Task-variant Score from Source Means}
For each source $s$, we compute the eigenspace mean
$\boldsymbol{\mu}_s=\frac{1}{N_s}\sum_{n:\mathrm{src}(n)=s}\mathbf{z}_n\in\mathbb{R}^{D}$.
Let $m$ be the component-wise median of $\{\mu_s\}_{s=1}^{S}$.
To ensure robustness against outlier datasets in multi-corpus settings, we quantify the task-variance of each dimension $j$ using a median-based dispersion metric (Eq.\ref{eq:reze-vj}), rather than the standard variance.
\begin{equation}
v_j=\frac{1}{S}\sum_{s=1}^{S}\big(\mu_{s,j}-m_j\big)^2,
\qquad j=1,\dots,D.
\label{eq:reze-vj}
\end{equation}

\paragraph{Active Dimensions}
We restrict suppression to a set of \emph{active} eigendimensions that explain most of variance.
Given a cumulative ratio $\rho$ (we use $0.99$), we select
\begin{equation}
k=\min\left\{k' : \frac{\sum_{j=1}^{k'}\lambda_j}{\sum_{j=1}^{D}\lambda_j}\ge \rho\right\}.
\label{eq:reze-active}
\end{equation}
Restricting the regularization to active dimensions prevents the inadvertent suppression of low-variance components that might contain fine-grained task-invariant signals, focusing instead on leading eigendimensions where biases are more pronounced.

\paragraph{Adaptive Soft-shrinkage Mechanism}
Let $\mathcal{A}=\{1,\dots,k\}$, and define a robust global threshold as follows.
Let $\tau$ be the median of $\{v_j\}_{j\in\mathcal{A}}$ and let $\mathrm{mad}$ be the median absolute deviation (MAD) of $\{v_j\}_{j\in\mathcal{A}}$.
Then
\begin{equation}
\theta=\tau+\gamma(\mathrm{mad}+\varepsilon),
\label{eq:reze-global-th}
\end{equation}
where $\gamma$ is a scale parameter(we use 1.0) and $\varepsilon>0$ is a small constant.
We additionally define a per-dimension band width using the MAD of source means:
\begin{equation}
\theta_j=\gamma\cdot \frac{1}{S}\sum_{s=1}^{S}|\mu_{s,j}-m_j|.
\label{eq:reze-band}
\end{equation}

For each source $s$ and dimension $j$, we set $\alpha_{s,j}=1$ by default, and update it only when (i) $j\in\mathcal{A}$ and $v_j>\theta$, and (ii) the source mean deviates beyond the band $|\Delta_{s,j}|\ge \theta_j$, where $\Delta_{s,j}=\mu_{s,j}-m_j$.
In that case, we set
\begin{equation}
\alpha_{s,j}
=
1+\eta\cdot \frac{m_j+\mathrm{sgn}(\Delta_{s,j})\theta_j-\mu_{s,j}}{|\mu_{s,j}|+\varepsilon},
\label{eq:reze-alpha}
\end{equation}
where $\eta$ controls the shrink strength. 
Unlike a hard-thresholding approach that zeros out dimensions \citep{mu2018all}, this soft shrinkage mechanism preserves the fundamental relational structure while selectively pulling task-specific deviations back toward the global consensus. 
In practice, we apply a simple clipping operation to ensure numerical stability and to avoid divergence or unintended reversal of the transformation.
Finally, we store $\mathbf{u}$, $\mathbf{W}$, and per-source diagonal shrink matrices $\mathbf{A}_s=\mathrm{diag}(\alpha_{s,1},\dots,\alpha_{s,D})$.
It is important to note that these parameters are pre-computed offline using the reference model $f_{0}$ before the pre-finetuning phase begins and remain fixed during training, incurring no additional computational overhead during the online update steps.

\subsection{Debiasing During Pre-finetuning}
\label{sec:online-reze-loss}

During pre-finetuning, for each training instance $i$ we denote its source identifier by $s_i$.
We debias the reference target relation $\mathbf{r}_{i}^{(0)}$ from the reference model $f_0$ using the source-specific shrinkage matrix $\mathbf{A}_{s_i}$:
\begin{equation}
\widehat{\mathbf{r}}_i^{(0)}
=
\mathbf{W}\mathbf{A}_{s_i}\mathbf{W}^\top\big(\mathbf{r}_i^{(0)}-\mathbf{u}\big)+\mathbf{u}.
\label{eq:reze-denoise}
\end{equation}

Then, we define the REZE regularization term $\mathcal{L}_{\mathrm{reze}}$ as the cosine dissimilarity between the current model relation $\mathbf{r}_i$ and the debiased target $\widehat{\mathbf{r}}_i^{(0)}$:
\begin{equation}
\mathcal{L}_{\mathrm{reze}} = \frac{1}{B}\sum_{i=1}^{B}\left(1-\cos\left(\mathbf{r}_i,\widehat{\mathbf{r}}_i^{(0)}\right)\right).
\label{eq:reze_regularization}
\end{equation}

Note that the debiased target $\widehat{\mathbf{r}}_i^{(0)}$ is derived from the fixed reference model parameters. Consequently, gradients are propagated solely through the current model’s relation prediction $\mathbf{r}_i$, acting as a directional guide.

\paragraph{Main objective}

Our model is optimized via the InfoNCE loss \citep{oord2018representation}, which contrasts positive pairs against in-batch negatives over a softmax distribution. This objective not only efficiently separates dissimilar instances but also creates a versatile embedding space compatible with classification and ranking paradigms. We adopt this standard objective to ensure robust performance across diverse downstream tasks such as retrieval and classification.

Specifically, let $\mathbf{e}^{a}_i$ and $\mathbf{e}^{p}_i$ denote the embeddings of the anchor and positive texts in a batch of size $B$, respectively.
Then

\begin{equation}
\mathcal{L}_{\mathrm{main}}
=
-\frac{1}{B}\sum_{i=1}^{B}
\log
\frac{
\exp\!\left(\mathrm{cos}(\mathbf{e}^{a}_i,\mathbf{e}^{p}_i)/\tau\right)
}{
\sum_{j=1}^{B}
\exp\!\left(\mathrm{cos}(\mathbf{e}^{a}_i,\mathbf{e}^{p}_j)/\tau\right)
},
\label{eq:reze-infonce}
\end{equation}
where $\tau$ is a temperature (we use 0.05).

\paragraph{Final objective}

The final objective combines the main loss $\mathcal{L}_{\mathrm{main}}$ with the REZE regularization term $\mathcal{L}_{\mathrm{reze}}$, where $\alpha$ is a hyperparameter that controls the strength of the regularization:
\begin{equation}
\mathcal{L} = \mathcal{L}_{\mathrm{main}} + \alpha \cdot \mathcal{L}_{\mathrm{reze}}.
\label{eq:reze-final}
\end{equation}

\paragraph{Batch Composition for Distribution Alignment}
REZE suppresses task-variant components in the eigenspace, encouraging the representation distributions induced by different corpora to become closer to one another. For this distribution alignment effect to operate effectively, the pre-finetuning stage requires batch compositions that adequately reflect heterogeneity across corpora and tasks. Accordingly, during REZE-based pre-finetuning, we impose no constraints on batch construction with respect to task or corpus, allowing examples from diverse sources to be mixed within each mini-batch.

\section{Experiments}

\subsection{Settings}
\label{subsec:settings}

To validate REZE in realistic domain adaptation regimes, we evaluate it under \textit{heterogeneous} pre-finetuning where multiple datasets with different task formats jointly shape the embedding space.
Following our two-stage pipeline, we (i) pre-finetune a general-purpose embedding model on heterogeneous in-domain tasks to acquire domain knowledge, and then (ii) fine-tune the model on a target task with limited supervision.
Hyperparameters and training environments across all experiments are reported in Appendix~\ref{app:hps}.

\subsubsection{Domains and Task Selection Protocol}
\label{subsec:domains-protocol}

\begin{table*}[t]
  \centering
  \small
  \setlength{\tabcolsep}{12pt}
  \begin{tabular}{l p{3.4cm} r r r }
    \hline
    Benchmark & Task-types & \# Target datasets & \# Datasets & \# Samples \\
    \hline

  FinMTEB
	& Classification & 3 & 5 & 17141 \\
    & PairClassification & - & 1 & 5285 \\
    & Reranking & 2 & 3 & 21744 \\
    & Retrieval & - & 8 & 58208 \\
    & STS & 1 & 1 & 35823 \\
  \hline
  Code (MTEB)
	& Retrieval & 3 & 6 & 172006 \\
  \hline
  ChemTEB
	& Classification & 3 & 17 & 107744 \\
    & Retrieval & 1 & 1 & 187 \\
    & Clustering & - & 2 & 2722 \\
    & PairClassification & - & 5 & 942 \\
  \hline

  \end{tabular}
  \caption{
  Summary of dataset composition per benchmark.
  }
  \label{tab:data_stats_summary}
\end{table*}

We evaluate REZE on three domain-specific benchmarks: Code(MTEB) \citep{muennighoff2023mteb}, ChemTEB \citep{pmlr-v262-shiraee-kasmaee24a}, and FinMTEB \citep{tang-yang-2025-finmteb}.
Each benchmark comprises multiple \textit{heterogeneous} task types (e.g., classification, STS, retrieval, reranking, and clustering), allowing us to assess whether pre-finetuning methods remain robust when supervision signals vary across tasks.
We intentionally choose these benchmarks to span diverse (i) \textbf{task-type coverage}, (ii) \textbf{dataset scale}, and (iii) \textbf{available training supervision}, thereby testing robustness across domains under realistic low-resource conditions.
For a fair comparison against fine-tuning (FT), we perform pre-finetuning using all available in-domain datasets except for the target dataset. We unify all pre-finetuning corpora into an anchor--positive pair format for contrastive training; detailed task-type-specific definitions are provided in Appendix~\ref{app:pair_construction}.

\paragraph{Task selection protocol}
For each benchmark, we first construct a candidate set of tasks by filtering to \textbf{English} tasks whose \textbf{train or test split} exceeds a minimum size threshold (Code(MTEB) and FinMTEB: $>1000$; ChemTEB: $>100$).
Among them, we define target-eligible tasks as those providing a \textbf{train split}.
From the resulting pool, we form a compact target set by selecting up to three tasks per task type within each benchmark.
We report the average score over the selected target tasks, and run three independent runs for \emph{all} settings using different random seeds for training-data shuffling. We report the per-task dataset statistics in Table~\ref{tab:data_stats_detailed}. 

\subsubsection{Models}
\label{subsec:models}
We evaluate REZE on multiple modern embedding backbones that differ in (i) model size($0.1\sim0.6$B), (ii) architecture and training recipe, (iii) training corpora, and (iv) model provider, to examine whether REZE is not tied to a specific embedding family.
Concretely, we choose E5 (0.1B) \citep{wang2022text}, ModernBERT (0.1B) \citep{warner2025smarter}, GTE (0.4B) \citep{zhang2024mgtegeneralizedlongcontexttext}, and Qwen3-Embedding (0.6B) \citep{zhang2025qwen3}.
For each model, we follow the recommended encoding configuration (pooling/normalization and any model-specific conventions).

\subsubsection{Baselines}
\label{subsec:baselines}

We compare the following training strategies. Unless stated otherwise, all methods share the same downstream fine-tuning protocol; the differences lie in whether and how we apply an additional pre-finetuning stage and/or evaluation-time post-processing.

\paragraph{FT}
Fine-tuning (FT) directly fits the base embedding model to the target task using only the target supervision, without any additional pre-finetuning stage.

\paragraph{PFT}
Pre-finetuning (PFT) first adapts the base model on heterogeneous in-domain data using the main contrastive objective (InfoNCE), and we then fine-tune the resulting model on the target task under the same downstream protocol as FT.
This baseline tests whether simply increasing in-domain exposure via heterogeneous pre-finetuning suffices, without explicitly controlling task-induced biases in the embedding space.

\paragraph{PFT$_{\text{Whitening}}$}
Whitening is a linear post-processing transformation that improves isotropy by re-centering and decorrelating representations \citep{su2021whitening}.
Concretely, given a set of sentence representations $\{\mathbf{h}_i\}_{i=1}^{M}$, we estimate the empirical mean $\boldsymbol{\mu}$ and covariance $\mathbf{C}$, and ensure isotropy by applying a whitening transform such as
$\mathbf{h}' = \mathbf{W}\mathbf{\Lambda}^{-\frac12}\mathbf{W}^\top(\mathbf{h}-\boldsymbol{\mu})$ where $\mathbf{C}=\mathbf{W}\mathbf{\Lambda}\mathbf{W}^\top$.
In our implementation, we follow the \emph{sequence-level} whitening pipeline provided by IsotropicIR \citep{jung2023isotropic}.
Specifically, after completing PFT and FT, we compute the whitening statistics on the target training set representations produced by the final fine-tuned encoder, and apply the resulting transform to all representations used during evaluation (queries/documents or sentence pairs, depending on the task).

\paragraph{PFT$_{\text{NormalizingFlow}}$}
Normalizing flows are expressive \emph{invertible} transformations that map a complex distribution to a simple reference distribution (e.g., a standard Gaussian) via a composition of bijective functions \citep{kobyzev2020normalizing}.
 \citet{jung2023isotropic} propose flow-based post-processing to enhance isotropy of encoder representations and provide a sequence-level implementation based on Glow \citep{kingma2018glow} (and NICE-style coupling \citep{dinh2014nice}).
In our experiment, we train a Glow-based sequence-level normalizing flow after completing PFT and FT, using the target training set representations produced by the final fine-tuned encoder (via maximum likelihood under the Gaussian reference).
We then transform all evaluation-time representations through the learned bijection before computing similarities for downstream evaluation.

\subsection{Result}

\begin{table*}[t]
  \centering
  \begin{tabular}{lccc ccc ccc}
    \hline
    & \multicolumn{3}{c}{FinMTEB} & \multicolumn{3}{c}{Code (MTEB)} & \multicolumn{3}{c}{ChemTEB} \\
    Method & 100 & 500 & 1000 & 100 & 500 & 1000 & 100 & 500 & 1000 \\
    \hline

    \multicolumn{10}{l}{\emph{E5 (e5-base-v2)}} \\
    FT
      & \underline{0.7202} & \underline{0.7986} & \underline{0.8125}
      & \underline{0.3906} & \underline{0.4800} & \underline{0.4898}
      & \textbf{0.8041} & \underline{0.8150} & \underline{0.6786} \\
    PFT
      & 0.6268 & 0.6793 & 0.7064
      & 0.3202 & 0.3386 & 0.3565
      & 0.7670 & 0.7902 & 0.6753 \\
    PFT$_{\text{Whitening}}$
      & 0.5158 & 0.5669 & 0.5686
      & 0.3249 & 0.3506 & 0.3630
      & 0.4375 & 0.4707 & 0.3953 \\
    PFT$_{\text{NormalizingFlow}}$
      & 0.5456 & 0.6003 & 0.6124
      & 0.2933 & 0.2974 & 0.3191
      & 0.5617 & 0.6112 & 0.5164 \\
    REZE
      & \textbf{0.7723} & \textbf{0.8050} & \textbf{0.8220}
      & \textbf{0.5002} & \textbf{0.5101} & \textbf{0.5286}
      & \underline{0.8007} & \textbf{0.8173} & \textbf{0.6833} \\
    \hline

    \multicolumn{10}{l}{\emph{ModernBERT (modernbert-embed-base)}} \\
    FT
      & \underline{0.7452} & \underline{0.8094} & \underline{0.8247}
      & \underline{0.5346} & \underline{0.5622} & \underline{0.5592}
      & \underline{0.8363} & \underline{0.8525} & 0.6837 \\
    PFT
      & 0.7048 & 0.7876 & 0.8192
      & 0.4394 & 0.5179 & 0.5158
      & 0.8241 & 0.8435 & \textbf{0.6870} \\
    PFT$_{\text{Whitening}}$
      & 0.5828 & 0.6418 & 0.6598
      & 0.4470 & 0.5074 & 0.5071
      & 0.4840 & 0.4775 & 0.3747 \\
    PFT$_{\text{NormalizingFlow}}$
      & 0.6283 & 0.7413 & 0.7457
      & 0.4224 & 0.4601 & 0.4666
      & 0.4357 & 0.4445 & 0.3521 \\
    REZE
      & \textbf{0.7929} & \textbf{0.8182} & \textbf{0.8373}
      & \textbf{0.5681} & \textbf{0.5652} & \textbf{0.5712}
      & \textbf{0.8515} & \textbf{0.8653} & \underline{0.6850} \\
    \hline

    \multicolumn{10}{l}{\emph{GTE (gte-large-en-v1.5)}} \\
    FT
      & 0.7150 & \underline{0.7260} & \underline{0.7584}
      & 0.5309 & 0.5239 & 0.5220
      & \underline{0.8358} & 0.8248 & 0.6668 \\
    PFT
      & \underline{0.7231} & 0.7191 & 0.7280
      & 0.5355 & 0.5352 & 0.5494
      & \textbf{0.8379} & \textbf{0.8443} & \textbf{0.6765} \\
    PFT$_{\text{Whitening}}$
      & 0.6737 & 0.6887 & 0.6534
      & \underline{0.5399} & \underline{0.5531} & \underline{0.5666}
      & 0.6694 & 0.5752 & 0.4369 \\
    PFT$_{\text{NormalizingFlow}}$
      & 0.7224 & 0.7230 & 0.7089
      & 0.4820 & 0.4811 & 0.5032
      & 0.7865 & 0.8013 & 0.6121 \\
    REZE
      & \textbf{0.7574} & \textbf{0.7752} & \textbf{0.7867}
      & \textbf{0.6189} & \textbf{0.6167} & \textbf{0.6206}
      & 0.8276 & \underline{0.8373} & \underline{0.6715} \\
    \hline

    \multicolumn{10}{l}{\emph{Qwen3-Embedding-0.6B}} \\
    FT
      & \underline{0.5697} & \underline{0.6898} & 0.6975
      & \underline{0.4019} & \textbf{0.4404} & \textbf{0.4632}
      & \underline{0.7537} & \textbf{0.7512} & 0.6563 \\
    PFT
      & 0.5545 & 0.6633 & \underline{0.7109}
      & 0.1214 & 0.1183 & 0.1819
      & 0.6481 & 0.6669 & \textbf{0.6765} \\
    PFT$_{\text{Whitening}}$
      & 0.3733 & 0.3557 & 0.3875
      & 0.1719 & 0.1900 & 0.2227
      & 0.4717 & 0.5328 & 0.5390 \\
    PFT$_{\text{NormalizingFlow}}$
      & 0.5121 & 0.5772 & 0.6013
      & 0.1447 & 0.1924 & 0.2108
      & 0.6793 & 0.6812 & \underline{0.6743} \\
    REZE
      & \textbf{0.7487} & \textbf{0.7288} & \textbf{0.7787}
      & \textbf{0.4081} & \underline{0.4291} & \underline{0.4556}
      & \textbf{0.7623} & \underline{0.7365} & 0.6688 \\
    \hline

  \end{tabular}
  \caption{
  Main results on three specialized benchmarks (FinMTEB, Code (MTEB), and ChemTEB) across different training sample sizes (100/500/1000).
  Each entry reports the \emph{average} score over the selected tasks within the corresponding benchmark.
  For each benchmark and training size, the best method is shown in \textbf{bold}, and the second-best method is \underline{underlined}.
  Detailed per-task results are provided in Tables~\ref{tab:per_task_results_1}--\ref{tab:per_task_results_5}.
  }
  \label{tab:main_results}
\end{table*}

\begin{figure*}[t]
  \centering
  \includegraphics[width=\textwidth]{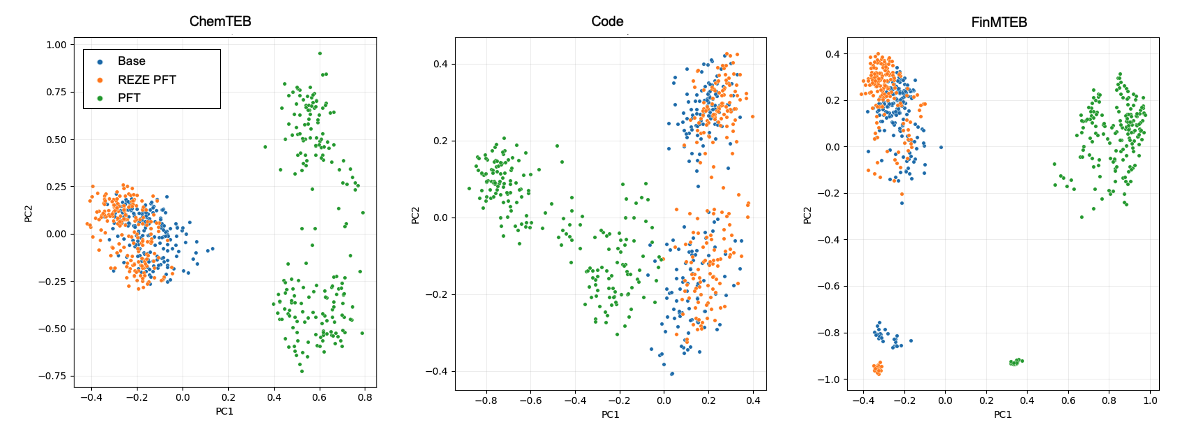}
  \caption{
    Embedding space visualization across three benchmarks.
    All datasets within each benchmark are encoded using the Qwen3 embedding model, and the resulting representations are projected to two dimensions using PCA ($n=2$).
    Each point corresponds to a single data instance represented by the concatenation of its anchor and positive embeddings, following the construction described in Section~\ref{sec:offline-reze-matrix}
    \emph{Base} denotes the original, non-trained embedding model.
  }
  \label{fig:embedding_form_analysis}
\end{figure*}

As shown in Table \ref{tab:main_results}, our method achieves superior performance across most settings. Notably, PFT—which injects domain knowledge through pre-finetuning—underperforms compared to FT, which directly fine-tunes on the target task. This finding provides direct empirical evidence that naively aggregating heterogeneous tasks can induce task conflicts and negative transfer. As illustrated in Figure \ref{fig:embedding_form_analysis}, REZE effectively preserves the original representation structure, whereas PFT distorts the representation geometry, thereby hindering transfer to downstream tasks. These results demonstrate that our representation regularization approach successfully absorbs shared domain knowledge while suppressing task-induced noise, yielding consistent performance improvements.

Meanwhile, performance degradation from PFT$_{\text{Whitening}}$ and PFT$_{\text{NormalizingFlow}}$ is also observed, particularly in the ChemTEB benchmark. We attribute this to the inherent post-processing principles of these methods: they manipulate embedding variance or transform representations into isotropic spaces based on already-fitted models. In specialized domains where general-purpose embedding models lack familiarity, such transformations distort spatial information.
Furthermore, the poor performance of PFT$_{\text{Whitening}}$ can also be attributed to estimating whitening statistics from highly limited data. When embedding distributions are insufficiently stabilized, statistics computed from few samples poorly represent the true distribution. Forcing unit variance under these conditions distorts useful structures, and the inverse rescaling of principal components excessively amplifies low-variance directions that are prone to underestimation, magnifying noise. Thus, global normalization based on limited training data disrupts the relative distance structure across tasks and domains, resulting in performance degradation. In contrast, our method proactively accounts for the noise originating from heterogeneous tasks during the pre-finetuning stage. This enables robust performance on downstream tasks and leads to consistent improvements.
 
\section{Analysis}

\begin{figure}[t]
  \includegraphics[width=\columnwidth]{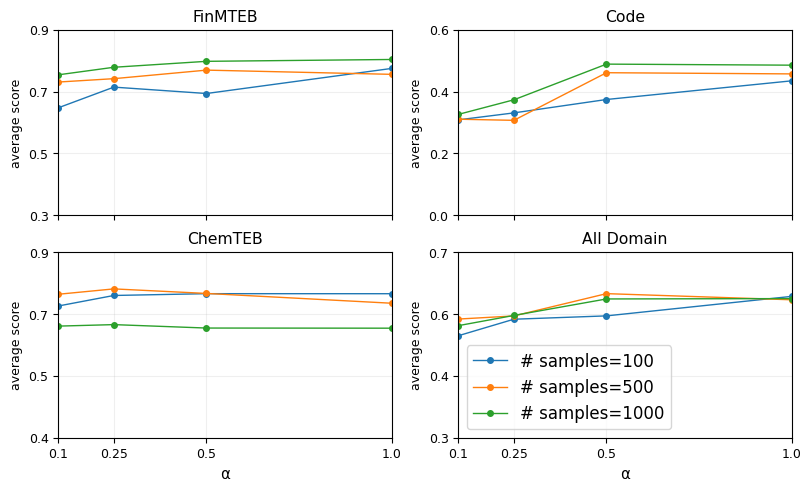}
  \caption{Effect of the regularization weight $\alpha$ on domain-average performance and their overall mean, across different training sample (\# samples = 100/500/1000), with the shrink strength fixed to $\eta=0.7$.}
  \label{fig:regularization_weight}
\end{figure}

\begin{table}[t]
  \centering
  \begin{tabular}{lcc}
    \hline
    Dataset & Mean & Median \\
    \hline
    ESGClassification & 0.8997 & 0.9117 \\
    FLSClassification & 0.8208 & 0.8205 \\
    FOMCClassification & 0.5446 & 0.5863 \\
    FiQA2018Reranking & 0.8420 & 0.9093 \\
    HC3Reranking & 0.9607 & 0.9598 \\
    FINAL & 0.5331 & 0.6172 \\
    \hline

  \end{tabular}
  \caption{
Comparison of mean and median aggregation on the FinMTEB benchmark using Qwen3-Embedding-0.6B.
  }
  \label{tab:mean_comparative_study}
\end{table}

\begin{figure}[t]
  \centering
  \includegraphics[width=0.8\columnwidth]{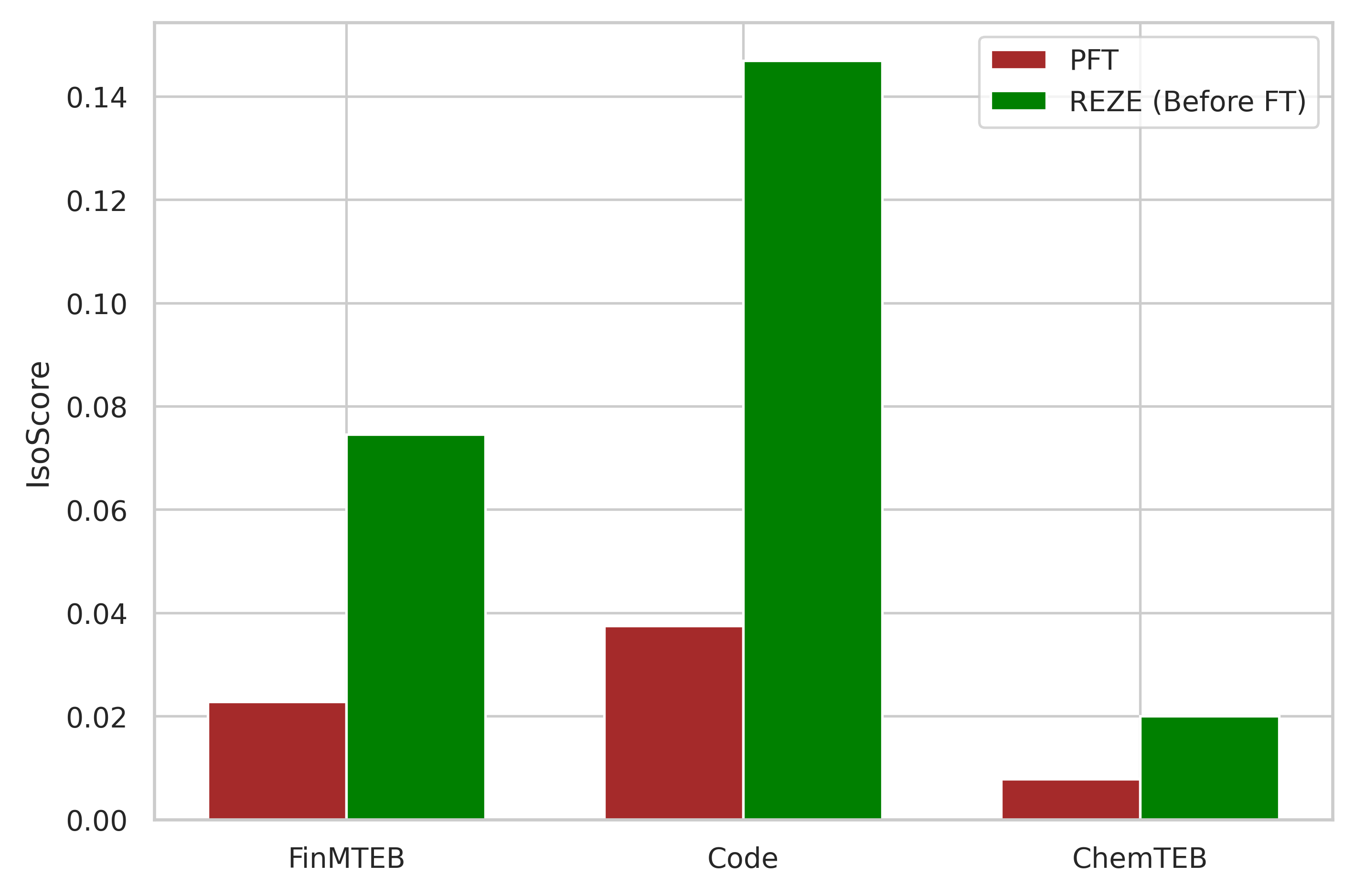}
  \caption{IsoScore comparison between PFT and REZE PFT across domains.}
  \label{fig:isotropy_analysis}
\end{figure}

\subsection{Regularization Weight Sweep}

The weight of regularization $\alpha$ determines how strongly the proposed regularization is applied during training, as defined in Eq.~\ref{eq:reze-final}. 
Figure~\ref{fig:regularization_weight} shows that increasing $\alpha$ generally improves average performance on FinMTEB and Code, while it produces inconsistent gains on ChemTEB. 
When aggregating results over all three domains, performance is most stable for $\alpha\ge0.5$, with $\alpha = 1.0$ achieving particularly strong results even in the low-data regime (\# samples = 100). 
This indicates that moderately strong regularization can improve sample efficiency without substantially harming overall performance. 
Based on these observations, we adopt $\alpha = 1.0$ as the default setting for the main experiments. 
In contrast, very large values of $\alpha$ (e.g., 5 or 10) tend to saturate or degrade performance on most tasks.

\subsection{Comparative Study: Mean vs. Median}

To validate the choice of median aggregation for estimating task-wise dispersion, we compare median and mean aggregation. Table~\ref{tab:mean_comparative_study} presents the task-level comparison between median and mean aggregation. The results demonstrate that median aggregation generally achieves higher scores than mean aggregation across the majority of tasks. This confirms that the median-based approach more effectively and robustly mitigates task-induced noise during pre-finetuning compared to standard mean-based statistics.

This performance gap stems from two primary theoretical considerations.
\textbf{First}, because representations are mean-centered before eigenspace projection, the global mean along each component is always zero. Using the mean as the shrinkage reference would therefore indiscriminately scale all values toward zero, destroying task-invariant semantic structures rather than selectively suppressing task-specific shifts.
\textbf{Second}, the mean is highly sensitive to outliers. Along active eigendimensions, representations frequently cluster by specific task characteristics. A mean-based reference can be disproportionately skewed by a small group of distant outliers, leading to unnecessary distortion of the informative embedding geometry. The median prevents this as a more robust estimator.

\subsection{Isotropy Analysis}

We examine whether the proposed regularization encourages a more isotropic embedding distribution in the target domains.
To this end, we measure isotropy with IsoScore~ \citep{rudman2022isoscore} for both standard pre-fine-tuning (PFT) and our method on each benchmark.
IsoScore quantifies how uniformly embedding dimensions are utilized; higher values indicate reduced concentration of variance along a small number of directions and thus a more isotropic distribution.
As shown in Figure~\ref{fig:isotropy_analysis}, our method consistently improves IsoScore over PFT across all benchmarks.
In particular, IsoScore increases by more than $\sim$3$\times$ on FinMTEB and Code, and by more than $\sim$2$\times$ on ChemTEB.
These results suggest that our regularization promotes more balanced use of embedding dimensions than standard PFT, forming an isotropic representation space.

\subsection{Representation Shift in Embedding Space}

Figure~\ref{fig:embedding_form_analysis} illustrates representation shifts under different pre-finetuning strategies by comparing the BASE model, PFT, and REZE across three benchmarks.

Contrastive objectives tend to push embeddings from an initially concentrated region toward a more dispersed configuration during fine-tuning~ \citep{ethayarajh2019contextual, xiao2023isotropy}.
Accordingly, PFT produces substantial drift in the global embedding distribution under heterogeneous task supervision.
While such expansion can be beneficial, our results suggest that uncontrolled drift may disrupt the pretrained geometry and, in extreme cases, lead to performance collapse for PFT variants (e.g., the Code (MTEB) scores of Qwen3 in Table~\ref{tab:main_results}).

In contrast, REZE exhibits a markedly different behavior.
Despite using the same contrastive main loss, REZE keeps embeddings closely aligned with the original geometry by constraining task-induced deviations, thereby suppressing task-specific noise without excessively distorting the pretrained manifold.
This controlled shift is crucial for pre-finetuning general-purpose embedding models, where preserving a well-structured similarity space is essential.
Consistent with both the visualization and downstream results, REZE is more robust to negative transfer, particularly in low-resource or highly heterogeneous pre-finetuning regimes.

\section{Conclusion}

We propose \textbf{REZE}, a representation regularization method that explicitly controls representation shift during pre-finetuning.
REZE operates on relation-level representations from anchor--positive pairs, decomposes them in an eigenspace, and detects task-variant directions with robust statistics, applying adaptive soft-shrinkage to suppress task-induced noise while preserving task-invariant semantics without inference-time overhead.
Designed for heterogeneous and low-resource pre-finetuning, REZE avoids the geometric distortion often introduced by isotropy-oriented post-processing such as whitening or normalizing flows.
Across multiple backbones and three specialized benchmarks, REZE consistently outperforms standard pre-finetuning and post-hoc regularization, and remains stable even in challenging cases such as Qwen3 on Code(MTEB) where PFT variants collapse.
Embedding-space analyses further confirm that REZE induces controlled shifts that stay aligned with the original manifold, suggesting that controlling representation shift—rather than enforcing isotropy—is key to robust embedding pre-finetuning under heterogeneous supervision.



\section*{Limitations}

Although we evaluate REZE on three widely used domain benchmarks (FinMTEB, Code(MTEB), and ChemTEB), the degree of domain specialization in publicly available embedding benchmarks can still be shallow compared to real-world professional settings.
In particular, for domains such as law—where domain-specific terminology is frequent, decisive, and often tied to jurisdictional nuances—there is a lack of openly accessible benchmarks that are both (i) sufficiently specialized and (ii) feasible to use for model training under the same experimental protocol.
Therefore, our results may underestimate the challenges (and potentially the benefits) of representation regularization in truly expert-level domains, and additional validation on such domains remains future work.

Due to limited computing resources, we could not conduct large-scale experiments on substantially larger embedding models, nor extensively explore training regimes that require much larger effective batch sizes.
In our experiments, we used a single NVIDIA A100 80GB GPU with a relatively small per-device batch size, which may affect the stability of contrastive learning and the behavior of in-batch negatives.
As a result, we do not claim that the observed trends directly extrapolate to much larger backbones or high-throughput training settings, and broader scaling studies (model size, batch size, and training duration) are left for future work.

\bibliography{custom} 

\appendix

\section{REZE Algorithm}
\label{app:algorithm}

This section provides the detailed pseudo-code for the REZE regularization workflow. 
As outlined in Algorithm~\ref{alg:reze}, the process is divided into two main phases: offline matrix construction and online debiasing.

\begin{algorithm}[t]
\caption{REZE Regularization Workflow}
\label{alg:reze}
\small
\KwIn{Source datasets $\{D_s\}_{s=1}^S$, reference model $f_0$, scale $\gamma$, strength $\eta$, variance ratio $\rho$}
\KwOut{Reze Matrix parameters $(\mathbf{u}, \mathbf{W}, \{\mathbf{A}_s\}_{s=1}^S)$}

\BlankLine
\tcp{Phase 1: Offline Reze Matrix Construction}
Collect relation vectors $\mathbf{x}_n = [f_0(x^a_n); f_0(x^p_n)]$ for all $n \in \bigcup D_s$\;
Compute global mean $\mathbf{u}$ and covariance $\mathbf{C} = \frac{1}{N}\tilde{\mathbf{X}}^\top\tilde{\mathbf{X}}$\;
Perform EVD: $\mathbf{C} = \mathbf{W}\mathbf{\Lambda}\mathbf{W}^\top$ and get z-space $\mathbf{z}_n = \mathbf{W}^\top(\mathbf{x}_n - \mathbf{u})$\;
Identify active dimensions $\mathcal{A}$ using $\rho$ per Eq.~\ref{eq:reze-active}\;

\For{each dimension $j \in \mathcal{A}$}{
    Compute source means $\mu_{s,j}$, median $m_j$, and variance $v_j$\;
    \If{$v_j > \theta$}{
        \For{each source $s=1 \dots S$}{
            \If{$|\mu_{s,j} - m_j| \ge \theta_j$}{
                Update $\alpha_{s,j}$ using soft shrinkage (Eq.~\ref{eq:reze-alpha})\;
            }
        }
    }
}

\BlankLine
\tcp{Phase 2: Online Debiasing during Pre-finetuning}
\For{each training iteration}{
    $\widehat{\mathbf{r}}_i^{(0)} \leftarrow \mathbf{W}\mathbf{A}_{t_i}\mathbf{W}^\top(\mathbf{r}_i^{(0)} - \mathbf{u}) + \mathbf{u}$ \tcp*{Debias target}
    $\mathcal{L} \leftarrow \mathcal{L}_{\mathrm{main}} + \alpha \cdot \mathcal{L}_{\mathrm{reze}}(\mathbf{r}_i, \widehat{\mathbf{r}}_i^{(0)})$ \tcp*{Update model}
}
\end{algorithm}

\section{Detailed Settings}
\label{app:hps}
Across all experiments, we use AdamW as the optimizer with a learning rate of $1\times 10^{-4}$ and a cosine learning-rate schedule.
We train with a per-device batch size of 8 for 1 epoch.
We truncate inputs to a maximum sequence length of 1024, except for E5 where we use its maximum positional encoding length of 512.
We train with \texttt{bfloat16} precision and run evaluation in \texttt{float32}.
All experiments are conducted on a single NVIDIA A100 80GB GPU.

\section{Data Details}
\label{app:data_details}

\subsection{Task-specific Preprocessing Procedure}
\label{app:pair_construction}

To enable pre-finetuning over heterogeneous datasets under a unified contrastive objective, we convert each dataset into an anchor--positive pair set.
Across all task types, we retain a pair only when the dataset provides an explicit positive semantic relation, and discard instances without such positive supervision.
We do not include explicit negative pairs in this preprocessing step.

\paragraph{Retrieval and Reranking}
For retrieval-style datasets, the \emph{anchor} is the query text and the \emph{positive} is a relevant document.
We retain only query--document pairs with positive relevance annotations.
For reranking datasets, each query is paired with its provided positive passages, and each (query, positive) pair is treated as an independent training instance.

\paragraph{Classification}
For classification datasets, the \emph{anchor} is the input text, and the \emph{positive} is the textual form of the ground-truth class label (i.e., the label name or label description when available).
This formulation injects label semantics into the embedding space while keeping the training format consistent.

\paragraph{Pair Classification}
For pair classification datasets, the \emph{anchor} and \emph{positive} are the two paired texts in the example.
We retain only positively labeled pairs (e.g., entailment/paraphrase/duplicate), and discard pairs without an explicit positive label.

\paragraph{Summarization}
For summarization datasets, the \emph{anchor} is the source document and the \emph{positive} is the corresponding reference summary.
We retain only examples that the dataset marks as valid (high-quality) document--summary alignments.

\paragraph{Clustering}
For clustering datasets, the \emph{anchor} is a set of sentences grouped by the dataset, and the \emph{positive} is the associated cluster label information.
This setup reflects group-level supervision rather than single-sentence pairing.

\paragraph{Summary}
Overall, pre-finetuning pairs are constructed in a positive-only manner: the anchor is the primary input text (query/document/sentence set), and the positive is a semantically aligned counterpart defined by task-specific supervision (relevant document, label text, paired sentence, or summary).

\subsection{Data Statistics}
\label{app:data_statistics}

We present detailed statistics of all datasets, including task types, evaluation metrics, input length distributions, and dataset sizes, as shown in Table~\ref{tab:data_stats_detailed}.

\begin{table*}[t]
  \centering
  \tiny
  \setlength{\tabcolsep}{5pt}
  \begin{tabular}{ll c ccc cccc r lc}
    \hline
    \textbf{Benchmark} & Task type & Metric
    & \multicolumn{4}{c}{Anchor length}
    & \multicolumn{4}{c}{Positive length}
    & \#Samples & \\ 
    \cline{4-7}\cline{8-11}
    Dataset & & & Avg & Min & Max & Med & Avg & Min & Max & Med & & Target \\
    \hline

    \multicolumn{13}{l}{\textbf{FinMTEB}} \\
    ESGClassification & Classification & accuracy & 168 & 35 & 1815 & 156 & 8 & 6 & 13 & 7 & 3,000 & \green{\text{\cmark}} \\
    FLSClassification & Classification & accuracy & 190 & 29 & 2210 & 170 & 10 & 7 & 16 & 7 & 2,600 & \green{\text{\cmark}} \\
	FOMCClassification & Classification & accuracy & 200 & 29 & 1249 & 183 & 6 & 6 & 7 & 7 & 1,281 & \green{\text{\cmark}} \\
    FiQA2018Reranking & Reranking & map & 62 & 14 & 166 & 61 & 1039 & 6 & 16984 & 788 & 13,511 & \green{\text{\cmark}} \\
	HC3Reranking & Reranking & map & 62 & 15 & 166 & 61 & 1105 & 11 & 10070 & 1084 & 5,866 & \green{\text{\cmark}} \\
    FINAL & STS & cosine\_spearman & 177 & 24 & 1821 & 158 & 177 & 24 & 1430 & 159 & 35,823 & \green{\text{\cmark}} \\
    
    FinancialPhraseBankClassification & Classification & accuracy & 121 & 9 & 315 & 108 & 7 & 7 & 8 & 7 & 1,264 & \red{\xmark} \\
	FinSentClassification & Classification & accuracy & 139 & 47 & 679 & 127 & 7 & 7 & 8 & 8 & 8,996 & \red{\xmark} \\
	HeadlineACPairClassification & PairClassification & ap & 50 & 6 & 140 & 50 & 50 & 10 & 117 & 50 & 5,285 & \red{\xmark} \\
	FinFactReranking & Reranking & map & 70 & 7 & 376 & 65 & 5476 & 396 & 46191 & 4685 & 2,367 & \red{\xmark} \\
	FiQA2018Retrieval & Retrieval & ndcg@10 & 62 & 14 & 166 & 61 & 1034 & 8 & 16986 & 786 & 17,072 & \red{\xmark} \\
	FinQARetrieval & Retrieval & ndcg@10 & 95 & 26 & 367 & 89 & 3994 & 228 & 16072 & 3954 & 8,281 & \red{\xmark} \\
	HC3Retrieval & Retrieval & ndcg@10 & 62 & 15 & 166 & 60 & 998 & 13 & 10072 & 739 & 3,933 & \red{\xmark} \\
	TATQARetrieval & Retrieval & ndcg@10 & 71 & 13 & 216 & 68 & 1812 & 245 & 9317 & 1344 & 1,668 & \red{\xmark} \\
	TheGoldmanEnRetrieval & Retrieval & ndcg@10 & 26 & 13 & 63 & 25 & 144 & 17 & 1591 & 121 & 1,512 & \red{\xmark} \\
	TradeTheEventEncyclopediaRetrieval & Retrieval & ndcg@10 & 29 & 10 & 117 & 27 & 4453 & 476 & 39836 & 3944 & 5,743 & \red{\xmark} \\
	USNewsRetrieval & Retrieval & ndcg@10 & 144 & 82 & 333 & 142 & 2518 & 8 & 112063 & 1376 & 9,999 & \red{\xmark} \\
	TradeTheEventNewsRetrieval & Retrieval & ndcg@10 & 217 & 74 & 1980 & 194 & 4773 & 227 & 94587 & 3735 & 10,000 & \red{\xmark} \\
    
    \hline
    \multicolumn{13}{l}{\textbf{Code (MTEB)}} \\
    CosQA & Retrieval & ndcg@10 & 37 & 16 & 95 & 35 & 278 & 87 & 6252 & 238 & 9,707 & \green{\text{\cmark}} \\
	StackOverflowQA & Retrieval & ndcg@10 & 1394 & 2 & 29809 & 882 & 1207 & 29 & 46027 & 726 & 13,951 & \green{\text{\cmark}} \\
	SyntheticText2SQL & Retrieval & ndcg@10 & 360 & 46 & 2143 & 340 & 127 & 16 & 761 & 111 & 100,000 & \green{\text{\cmark}} \\
    
	AppsRetrieval & Retrieval & ndcg@10 & 1669 & 152 & 5742 & 1601 & 586 & 14 & 14348 & 398 & 3,765 & \red{\xmark} \\
	CodeFeedbackMT & Retrieval & ndcg@10 & 4425 & 127 & 32432 & 3568 & 1478 & 1 & 7830 & 1376 & 13,277 & \red{\xmark} \\
	CodeFeedbackST & Retrieval & ndcg@10 & 724 & 26 & 13849 & 539 & 1525 & 1 & 10042 & 1399 & 31,306 & \red{\xmark} \\
    \hline

    \multicolumn{13}{l}{\textbf{ChemTEB}} \\
    WikipediaCryobiologySeparationClassification & Classification & accuracy & 1098 & 105 & 19539 & 861 & 16 & 11 & 21 & 17 & 931 & \green{\text{\cmark}} \\
    WikipediaTheoreticalAppliedClassification & Classification & accuracy & 993 & 105 & 4574 & 763 & 37 & 37 & 39 & 37 & 46,661 & \green{\text{\cmark}} \\
	WikipediaCrystallographyAnalyticalClassification & Classification & accuracy & 1169 & 107 & 10968 & 852 & 27 & 15 & 42 & 15 & 1,160 & \green{\text{\cmark}} \\
	ChemHotpotQARetrieval & Retrieval & ndcg@10 & 116 & 32 & 481 & 102 & 460 & 81 & 1352 & 398 & 187 & \green{\text{\cmark}} \\
    
	SDSEyeProtectionClassification & Classification & accuracy & 3493 & 789 & 11292 & 3371 & 25 & 25 & 29 & 25 & 6,000 & \red{\xmark} \\
	SDSGlovesClassification & Classification & accuracy & 3522 & 307 & 13127 & 3372 & 16 & 16 & 20 & 16 & 6,000 & \red{\xmark} \\
	WikipediaBioMetChemClassification & Classification & accuracy & 1103 & 105 & 13719 & 822 & 22 & 10 & 37 & 10 & 4,592 & \red{\xmark} \\
	WikipediaGreenhouseEnantiopureClassification & Classification & accuracy & 831 & 106 & 11179 & 613 & 16 & 16 & 17 & 17 & 908 & \red{\xmark} \\
	WikipediaSolidStateColloidalClassification & Classification & accuracy & 978 & 105 & 18971 & 701 & 20 & 19 & 21 & 21 & 1,772 & \red{\xmark} \\
	WikipediaOrganicInorganicClassification & Classification & accuracy & 773 & 105 & 6504 & 576 & 27 & 17 & 41 & 17 & 1,049 & \red{\xmark} \\
	WikipediaChemistryTopicsClassification & Classification & accuracy & 1118 & 105 & 19539 & 853 & 15 & 11 & 21 & 14 & 1,684 & \red{\xmark} \\
	WikipediaChemFieldsClassification & Classification & accuracy & 985 & 105 & 4438 & 761 & 15 & 10 & 24 & 15 & 34,173 & \red{\xmark} \\
	WikipediaLuminescenceClassification & Classification & accuracy & 969 & 127 & 5804 & 694 & 13 & 12 & 15 & 15 & 328 & \red{\xmark} \\
	WikipediaIsotopesFissionClassification & Classification & accuracy & 1256 & 118 & 6585 & 982 & 20 & 8 & 34 & 8 & 333 & \red{\xmark} \\
	WikipediaSaltsSemiconductorsClassification & Classification & accuracy & 811 & 106 & 5279 & 629 & 13 & 5 & 23 & 5 & 392 & \red{\xmark} \\
	WikipediaBiolumNeurochemClassification & Classification & accuracy & 1012 & 106 & 5804 & 748 & 14 & 14 & 15 & 14 & 388 & \red{\xmark} \\
	WikipediaCompChemSpectroscopyClassification & Classification & accuracy & 1073 & 105 & 10922 & 726 & 17 & 15 & 22 & 15 & 880 & \red{\xmark} \\
	WikipediaChemEngSpecialtiesClassification & Classification & accuracy & 946 & 108 & 6844 & 651 & 15 & 11 & 20 & 15 & 493 & \red{\xmark} \\
	WikipediaChemistryTopicsClustering & Clustering & v\_measure & 1129 & 105 & 19539 & 862 & 15 & 11 & 21 & 14 & 2,105 & \red{\xmark} \\
	WikipediaSpecialtiesInChemistryClustering & Clustering & v\_measure & 906 & 108 & 6844 & 619 & 15 & 11 & 20 & 15 & 617 & \red{\xmark} \\
	PubChemAISentenceParaphrasePC & PairClassification & max\_ap & 86 & 23 & 420 & 67 & 104 & 28 & 433 & 88 & 205 & \red{\xmark} \\
	PubChemSMILESPC & PairClassification & max\_ap & 86 & 5 & 694 & 35 & 46 & 19 & 100 & 44 & 204 & \red{\xmark} \\
	PubChemSynonymPC & PairClassification & max\_ap & 16 & 4 & 86 & 14 & 35 & 5 & 175 & 26 & 204 & \red{\xmark} \\
	PubChemWikiParagraphsPC & PairClassification & max\_ap & 353 & 22 & 2094 & 242 & 383 & 16 & 5046 & 255 & 205 & \red{\xmark} \\
	PubChemWikiPairClassification & PairClassification & max\_ap & 315 & 25 & 1281 & 223 & 585 & 70 & 2609 & 478 & 124 & \red{\xmark} \\
    \hline

  \end{tabular}
  \caption{
  Detailed dataset statistics and metrics for each benchmark.
  }
  \label{tab:data_stats_detailed}
\end{table*}

\FloatBarrier

\section{Per-task Result}
\label{app:per_task_result}

\begin{table*}[t]
  \centering
  \begin{tabular}{lccc ccc ccc}
    \hline
    FinMTEB & \multicolumn{3}{c}{ESGClassification} & \multicolumn{3}{c}{FLSClassification} & \multicolumn{3}{c}{FOMCClassification} \\
    Method & 100 & 500 & 1000 & 100 & 500 & 1000 & 100 & 500 & 1000 \\
    \hline

    \multicolumn{10}{l}{\emph{E5 (e5-base-v2)}} \\
    FT
	   & 0.8421 & 0.9242 & 0.9334
	   & 0.7080 & 0.8338 & 0.8485
	   & 0.4496 & 0.5845 & 0.6098 \\
    PFT
	   & 0.7697 & 0.9038 & 0.9119
	   & 0.4053 & 0.4005 & 0.4125
	   & 0.4853 & 0.5649 & 0.5898 \\
    PFT$_{\text{Whitening}}$
	   & 0.4089 & 0.5712 & 0.5055
	   & 0.3900 & 0.3799 & 0.3831
	   & 0.3466 & 0.3858 & 0.3846 \\
    PFT$_{\text{NormalizingFlow}}$
	  & 0.4328 & 0.6056 & 0.5466
	  & 0.4102 & 0.4076 & 0.4079
	  & 0.3721 & 0.4092 & 0.4106 \\
    REZE
	   & 0.9047 & 0.9327 & 0.9358
	   & 0.7428 & 0.8329 & 0.8470
	   & 0.5198 & 0.5567 & 0.6099 \\
    \hline

    \multicolumn{10}{l}{\emph{ModernBERT (modernbert-embed-base)}} \\
    FT
	   & 0.8744 & 0.9119 & 0.9289
	   & 0.6722 & 0.8197 & 0.8264
	   & 0.5101 & 0.5942 & 0.6441 \\
    PFT
	   & 0.8683 & 0.8931 & 0.9302
	   & 0.5440 & 0.8186 & 0.8523
	   & 0.5041 & 0.5710 & 0.5940 \\
    PFT$_{\text{Whitening}}$
	   & 0.4463 & 0.5580 & 0.5783
	   & 0.4032 & 0.5346 & 0.5932
	   & 0.3699 & 0.4022 & 0.3998 \\
    PFT$_{\text{NormalizingFlow}}$
	  & 0.4814 & 0.6580 & 0.6656
	  & 0.4322 & 0.6353 & 0.6762
	  & 0.4046 & 0.4779 & 0.4586 \\
    REZE
	   & 0.9059 & 0.9323 & 0.9389
	   & 0.7831 & 0.8094 & 0.8427
	   & 0.5700 & 0.5972 & 0.6547 \\
    \hline

    \multicolumn{10}{l}{\emph{GTE (gte-large-en-v1.5)}} \\
    FT
	  & 0.8914 & 0.9194 & 0.9111
	  & 0.6135 & 0.4678 & 0.5697
	  & 0.4256 & 0.4822 & 0.5483 \\
    PFT
	  & 0.8550 & 0.8774 & 0.9153
	  & 0.7439 & 0.4287 & 0.3853
	  & 0.4488 & 0.5159 & 0.5267 \\
    PFT$_{\text{Whitening}}$
	  & 0.6562 & 0.8345 & 0.6496
	  & 0.6599 & 0.4367 & 0.3446
	  & 0.4567 & 0.4853 & 0.4859 \\
    PFT$_{\text{NormalizingFlow}}$
	  & 0.6120 & 0.9041 & 0.7590
	  & 0.6149 & 0.4754 & 0.4137
	  & 0.4269 & 0.5273 & 0.5710 \\
    REZE
	  & 0.9034 & 0.9079 & 0.9235
	  & 0.6904 & 0.7579 & 0.8024
	  & 0.4859 & 0.4851 & 0.5030 \\
    \hline

    \multicolumn{10}{l}{\emph{Qwen3-Embedding-0.6B}} \\
    FT
	  & 0.7878 & 0.8964 & 0.8541
	  & 0.3412 & 0.3668 & 0.4226
	  & 0.3542 & 0.4821 & 0.5170 \\
    PFT
	  & 0.6924 & 0.8014 & 0.8486
	  & 0.3745 & 0.6316 & 0.7139
	  & 0.3933 & 0.4678 & 0.5162 \\
    PFT$_{\text{Whitening}}$
	  & 0.4556 & 0.6325 & 0.6733
	  & 0.4193 & 0.5446 & 0.6314
	  & 0.4212 & 0.5331 & 0.6010 \\
    PFT$_{\text{NormalizingFlow}}$
	  & 0.5356 & 0.6671 & 0.6411
	  & 0.4885 & 0.4872 & 0.5614
	  & 0.4701 & 0.4434 & 0.5182 \\
    REZE
	  & 0.8620 & 0.9061 & 0.9117
	  & 0.7292 & 0.8155 & 0.8205
	  & 0.4786 & 0.5519 & 0.5863 \\
    \hline   
  \end{tabular}
  \caption{
  Main results (1) on FinMTEB across different training sample sizes (100/500/1000).
  Each entry reports the score over the selected tasks within the corresponding benchmark.
  }
  \label{tab:per_task_results_1}
\end{table*}

\begin{table*}[t]
  \centering
  \begin{tabular}{lccc ccc ccc}
    \hline
    FinMTEB & \multicolumn{3}{c}{FiQA2018Reranking} & \multicolumn{3}{c}{HC3Reranking} & \multicolumn{3}{c}{FINAL} \\
    Method & 100 & 500 & 1000 & 100 & 500 & 1000 & 100 & 500 & 1000 \\
    \hline

    \multicolumn{10}{l}{\emph{E5 (e5-base-v2)}} \\
    FT
	   & 0.8711 & 0.9304 & 0.9438
	   & 0.9429 & 0.9603 & 0.9654
	   & 0.5077 & 0.5586 & 0.5740 \\
    PFT
	   & 0.8340 & 0.8721 & 0.8896
	   & 0.9267 & 0.9448 & 0.9517
	   & 0.3396 & 0.3895 & 0.4831 \\
    PFT$_{\text{Whitening}}$
	   & 0.6981 & 0.7585 & 0.7659
	   & 0.8826 & 0.8801 & 0.8872
	   & 0.3685 & 0.4260 & 0.4853 \\
    PFT$_{\text{NormalizingFlow}}$
	  & 0.7385 & 0.7881 & 0.8019
	  & 0.9272 & 0.9381 & 0.9731
	  & 0.3925 & 0.4528 & 0.5340 \\
    REZE
	   & 0.9312 & 0.9272 & 0.9389
	   & 0.9763 & 0.9735 & 0.9720
	   & 0.5588 & 0.6073 & 0.6283 \\
    \hline

    \multicolumn{10}{l}{\emph{ModernBERT (modernbert-embed-base)}} \\
    FT
	   & 0.9538 & 0.9543 & 0.9621
	   & 0.9735 & 0.9774 & 0.9741
	   & 0.4871 & 0.5989 & 0.6128 \\
    PFT
	   & 0.9264 & 0.9354 & 0.9441
	   & 0.9479 & 0.9722 & 0.9759
	   & 0.4383 & 0.5351 & 0.6187 \\
    PFT$_{\text{Whitening}}$
	   & 0.8188 & 0.8410 & 0.8519
	   & 0.9287 & 0.9269 & 0.9242
	   & 0.5300 & 0.5883 & 0.6117 \\
    PFT$_{\text{NormalizingFlow}}$
	  & 0.8827 & 0.9697 & 0.9516
	  & 0.9944 & 0.9813 & 0.9766
	  & 0.5742 & 0.6939 & 0.7118 \\
    REZE
	   & 0.9552 & 0.9594 & 0.9653
	   & 0.9869 & 0.9850 & 0.9872
	   & 0.5561 & 0.6259 & 0.6350 \\
    \hline

    \multicolumn{10}{l}{\emph{GTE (gte-large-en-v1.5)}} \\
    FT
	  & 0.9408 & 0.9248 & 0.9401
	  & 0.9747 & 0.9753 & 0.9782
	  & 0.4438 & 0.5862 & 0.6031 \\
    PFT
	  & 0.9349 & 0.9519 & 0.9609
	  & 0.9608 & 0.9745 & 0.9737
	  & 0.3950 & 0.5662 & 0.6061 \\
    PFT$_{\text{Whitening}}$
	  & 0.8900 & 0.8790 & 0.8880
	  & 0.9546 & 0.9568 & 0.9632
	  & 0.4249 & 0.5398 & 0.5893 \\
    PFT$_{\text{NormalizingFlow}}$
	  & 0.8300 & 0.9500 & 0.9934
	  & 0.8890 & 0.9914 & 0.9893
	  & 0.3967 & 0.5853 & 0.6924 \\
    REZE
	  & 0.9553 & 0.9456 & 0.9506
	  & 0.9731 & 0.9811 & 0.9811
	  & 0.5362 & 0.5737 & 0.5593 \\
    \hline

    \multicolumn{10}{l}{\emph{Qwen3-Embedding-0.6B}} \\
    FT
	  & 0.7567 & 0.8771 & 0.9038
	  & 0.8890 & 0.9148 & 0.9382
	  & 0.2892 & 0.6019 & 0.5494 \\
    PFT
	  & 0.7270 & 0.7358 & 0.7840
	  & 0.7618 & 0.8356 & 0.8480
	  & 0.3780 & 0.5078 & 0.5547 \\
    PFT$_{\text{Whitening}}$
	  & 0.4606 & 0.4832 & 0.4981
	  & 0.5440 & 0.5419 & 0.5784
	  & 0.3602 & 0.4742 & 0.5222 \\
    PFT$_{\text{NormalizingFlow}}$
	  & 0.4912 & 0.5013 & 0.5412
	  & 0.7738 & 0.8523 & 0.8661
	  & 0.3651 & 0.5123 & 0.5374 \\
    REZE
	  & 0.8949 & 0.9033 & 0.9093
	  & 0.9581 & 0.9503 & 0.9598
	  & 0.4344 & 0.6324 & 0.6172 \\
    \hline
  \end{tabular}
  \caption{
  Main results (2) on FinMTEB across different training sample sizes (100/500/1000).
  Each entry reports the score over the selected tasks within the corresponding benchmark.
  }
  \label{tab:per_task_results_2}
\end{table*}

\begin{table*}[t]
  \centering
  \begin{tabular}{lccc ccc ccc}
    \hline
    Code & \multicolumn{3}{c}{CosQA} & \multicolumn{3}{c}{StackOverflowQA} & \multicolumn{3}{c}{SyntheticText2SQL} \\
    Method & 100 & 500 & 1000 & 100 & 500 & 1000 & 100 & 500 & 1000 \\
    \hline

    \multicolumn{10}{l}{\emph{E5 (e5-base-v2)}} \\
    FT
	   & 0.2610 & 0.2495 & 0.2619
	   & 0.5801 & 0.7198 & 0.7312
	   & 0.3308 & 0.4707 & 0.4763 \\
    PFT
	   & 0.1714 & 0.1744 & 0.1875
	   & 0.3907 & 0.3992 & 0.4445
	   & 0.3986 & 0.4422 & 0.4374 \\
    PFT$_{\text{Whitening}}$
	   & 0.1596 & 0.1623 & 0.1690
	   & 0.4540 & 0.4749 & 0.5118
	   & 0.3610 & 0.4147 & 0.4082 \\
    PFT$_{\text{NormalizingFlow}}$
	  & 0.1435 & 0.1366 & 0.1517
	  & 0.4077 & 0.4049 & 0.4479
	  & 0.3285 & 0.3505 & 0.3576 \\
    REZE
	   & 0.2693 & 0.3037 & 0.3110
	   & 0.7289 & 0.7343 & 0.7453
	   & 0.5023 & 0.4922 & 0.5296 \\
    \hline

    \multicolumn{10}{l}{\emph{ModernBERT (modernbert-embed-base)}} \\
    FT
	   & 0.3183 & 0.3378 & 0.3400
	   & 0.7466 & 0.7965 & 0.7910
	   & 0.5390 & 0.5523 & 0.5465 \\
    PFT
	   & 0.2417 & 0.3093 & 0.3118
	   & 0.6546 & 0.7298 & 0.7405
	   & 0.4218 & 0.5146 & 0.4952 \\
    PFT$_{\text{Whitening}}$
	   & 0.2051 & 0.2576 & 0.2628
	   & 0.7248 & 0.7915 & 0.7996
	   & 0.4109 & 0.4733 & 0.4588 \\
    PFT$_{\text{NormalizingFlow}}$
	  & 0.1929 & 0.2318 & 0.2466
	  & 0.6813 & 0.7211 & 0.7327
	  & 0.3929 & 0.4272 & 0.4203 \\
    REZE
	   & 0.3381 & 0.3447 & 0.3461
	   & 0.8193 & 0.7981 & 0.8132
	   & 0.5469 & 0.5528 & 0.5543 \\
    \hline

    \multicolumn{10}{l}{\emph{GTE (gte-large-en-v1.5)}} \\
    FT
	  & 0.2765 & 0.2725 & 0.3034
	  & 0.7797 & 0.7703 & 0.7480
	  & 0.5364 & 0.5290 & 0.5147 \\
    PFT
	  & 0.3053 & 0.2866 & 0.3130
	  & 0.7600 & 0.7599 & 0.7683
	  & 0.5413 & 0.5592 & 0.5669 \\
    PFT$_{\text{Whitening}}$
	  & 0.2865 & 0.2844 & 0.2970
	  & 0.7938 & 0.8153 & 0.8305
	  & 0.5393 & 0.5598 & 0.5722 \\
    PFT$_{\text{NormalizingFlow}}$
	  & 0.2239 & 0.2370 & 0.2440
	  & 0.6203 & 0.6781 & 0.6862
	  & 0.4228 & 0.4651 & 0.4695 \\
    REZE
	  & 0.3888 & 0.3751 & 0.3825
	  & 0.8872 & 0.8838 & 0.8841
	  & 0.5806 & 0.5913 & 0.5945 \\
    \hline

    \multicolumn{10}{l}{\emph{Qwen3-Embedding-0.6B}} \\
    FT
	  & 0.1058 & 0.0950 & 0.1855
	  & 0.7228 & 0.7058 & 0.7227
	  & 0.3770 & 0.5205 & 0.4815 \\
    PFT
	  & 0.0669 & 0.0515 & 0.0225
	  & 0.2678 & 0.2017 & 0.3704
	  & 0.0296 & 0.1016 & 0.1527 \\
    PFT$_{\text{Whitening}}$
	  & 0.0450 & 0.0745 & 0.0565
	  & 0.4435 & 0.4426 & 0.5171
	  & 0.0272 & 0.0530 & 0.0946 \\
    PFT$_{\text{NormalizingFlow}}$
	  & 0.0214 & 0.2310 & 0.2713
	  & 0.3073 & 0.2226 & 0.2986
	  & 0.1053 & 0.1235 & 0.0623 \\
    REZE
	  & 0.2309 & 0.2637 & 0.2648
	  & 0.7908 & 0.7627 & 0.7769
	  & 0.4900 & 0.4682 & 0.5332 \\
    \hline
  \end{tabular}
  \caption{
  Main results on Code (MTEB) across different training sample sizes (100/500/1000).
  Each entry reports the score over the selected tasks within the corresponding benchmark.
  }
  \label{tab:per_task_results_3}
\end{table*}

\begin{table*}[t]
  \centering
  \begin{tabular}{lccc ccc ccc}
    \hline
    ChemTEB & \multicolumn{3}{c}{CryobiologySeparation} & \multicolumn{3}{c}{TheoreticalApplied} & \multicolumn{3}{c}{ChemHotpotQARetrieval} \\
    Method & 100 & 500 & 1000 & 100 & 500 & 1000 & 100 & 500 & 1000 \\
    \hline

    \multicolumn{10}{l}{\emph{E5 (e5-base-v2)}} \\
    FT
	   & 0.9416 & 0.9421 & 0.9519
	   & 0.7286 & 0.7653 & 0.7722
	  & 0.5558 & 0.5590 & -- \\
    PFT
	   & 0.9185 & 0.9356 & 0.9446
	   & 0.6887 & 0.7414 & 0.7677
	  & 0.4882 & 0.4980 & -- \\
    PFT$_{\text{Whitening}}$
	   & 0.3060 & 0.2794 & 0.2970
	   & 0.5254 & 0.6459 & 0.6015
	  & 0.3573 & 0.3573 & -- \\
    PFT$_{\text{NormalizingFlow}}$
	  & 0.3925 & 0.3650 & 0.2946
	  & 0.6727 & 0.8446 & 0.5870
	  & 0.4582 & 0.4533 & -- \\
    REZE
	   & 0.9554 & 0.9455 & 0.9592
	   & 0.7353 & 0.7758 & 0.7842
	  & 0.5445 & 0.5590 & -- \\
    \hline

    \multicolumn{10}{l}{\emph{ModernBERT (modernbert-embed-base)}} \\
    FT
	   & 0.9528 & 0.9605 & 0.9695
	   & 0.7306 & 0.7554 & 0.7745
	  & 0.6763 & 0.7004 & -- \\
    PFT
	   & 0.9532 & 0.9631 & 0.9747
	   & 0.7119 & 0.7716 & 0.7824
	  & 0.6510 & 0.6506 & -- \\
    PFT$_{\text{Whitening}}$
 	   & 0.3403 & 0.3433 & 0.3395
 	   & 0.5346 & 0.5497 & 0.5448
	  & 0.3684 & 0.4128 & -- \\
    PFT$_{\text{NormalizingFlow}}$
	  & 0.3059 & 0.3211 & 0.2418
	  & 0.4789 & 0.5157 & 0.3823
	  & 0.3310 & 0.3772 & -- \\
    REZE
	   & 0.9562 & 0.9614 & 0.9648
	   & 0.7281 & 0.7781 & 0.7905
	  & 0.7295 & 0.7344 & --\\
    \hline

    \multicolumn{10}{l}{\emph{GTE (gte-large-en-v1.5)}} \\
    FT
	  & 0.9687 & 0.9554 & 0.9459
	  & 0.7118 & 0.7107 & 0.7243
	  & 0.6701 & 0.6385 & -- \\
    PFT
	  & 0.9652 & 0.9652 & 0.9751
	  & 0.7124 & 0.7287 & 0.7321
	  & 0.6857 & 0.6907 & -- \\
    PFT$_{\text{Whitening}}$
	  & 0.7519 & 0.2961 & 0.2845
	  & 0.5950 & 0.7281 & 0.7216
	  & 0.4684 & 0.4128 & -- \\
    PFT$_{\text{NormalizingFlow}}$
	  & 0.4892 & 0.2298 & 0.1714
	  & 0.3867 & 0.5631 & 0.4383
	  & 0.3047 & 0.3177 & -- \\
    REZE
	  & 0.9489 & 0.9712 & 0.9627
	  & 0.6818 & 0.7427 & 0.7462
	  & 0.6940 & 0.6686 & -- \\
    \hline

    \multicolumn{10}{l}{\emph{Qwen3-Embedding-0.6B}} \\
    FT
	  & 0.9189 & 0.9292 & 0.9528
	  & 0.6138 & 0.7138 & 0.7599
	  & 0.5775 & 0.4827 & -- \\
    PFT
	  & 0.8966 & 0.9279 & 0.9532
	  & 0.7218 & 0.7280 & 0.7629
	  & 0.0030 & 0.0215 & -- \\
    PFT$_{\text{Whitening}}$
	  & 0.4803 & 0.6288 & 0.5871
	  & 0.6074 & 0.6895 & 0.7426
	  & 0.0350 & 0.0008 & -- \\
    PFT$_{\text{NormalizingFlow}}$
	  & 0.7168 & 0.8158 & 0.5506
	  & 0.8993 & 0.8912 & 0.6970
	  & 0.0288 & 0.2695 & -- \\
    REZE
	  & 0.9227 & 0.9567 & 0.9622
	  & 0.7225 & 0.7604 & 0.7738
	  & 0.5437 & 0.5741 & -- \\
    \hline
  \end{tabular}
  \caption{
  Main results (1) on ChemTEB across different training sample sizes (100/500/1000).
  Each entry reports the score over the selected tasks within the corresponding benchmark.
  }
  \label{tab:per_task_results_4}
\end{table*}

\begin{table*}[t]
  \centering
  \begin{tabular}{lccc ccc ccc}
    \hline
    ChemTEB & \multicolumn{3}{c}{CrystallographyAnalytical} & \multicolumn{3}{c}{ -- } & \multicolumn{3}{c}{ -- } \\
    Method & 100 & 500 & 1000 & 100 & 500 & 1000 & 100 & 500 & 1000 \\
    \hline

    \multicolumn{10}{l}{\emph{E5 (e5-base-v2)}} \\
    FT
	   & 0.9904 & 0.9935 & 0.9904
	  & -- & -- & --
	  & -- & -- & -- \\
    PFT
	   & 0.9725 & 0.9859 & 0.9890
	  & -- & -- & --
	  & -- & -- & -- \\
    PFT$_{\text{Whitening}}$
	   & 0.5615 & 0.6003 & 0.6828
	  & -- & -- & --
	  & -- & -- & -- \\
    PFT$_{\text{NormalizingFlow}}$
	   & 0.7233 & 0.7817 & 0.6675
	  & -- & -- & --
	  & -- & -- & -- \\
    REZE
	   & 0.9677 & 0.9890 & 0.9897 
	  & -- & -- & --
	  & -- & -- & -- \\
    \hline

    \multicolumn{10}{l}{\emph{ModernBERT (modernbert-embed-base)}} \\
    FT
	   & 0.9801 & 0.9887 & 0.9911
	  & -- & -- & --
	  & -- & -- & -- \\
    PFT
	   & 0.9801 & 0.9887 & 0.9911
	  & -- & -- & --
	  & -- & -- & -- \\
    PFT$_{\text{Whitening}}$
  	 & 0.6928 & 0.6041 & 0.6144
	  & -- & -- & --
	  & -- & -- & -- \\
    PFT$_{\text{NormalizingFlow}}$
	  & 0.6268 & 0.5638 & 0.4320
	  & -- & -- & --
	  & -- & -- & -- \\
    REZE
	   & 0.9921 & 0.9873 & 0.9849
	  & -- & -- & --
	  & -- & -- & -- \\
    \hline

    \multicolumn{10}{l}{\emph{GTE (gte-large-en-v1.5)}} \\
    FT
	  & 0.9928 & 0.9945 & 0.9969
	  & -- & -- & --
	  & -- & -- & -- \\
    PFT
	  & 0.9883 & 0.9924 & 0.9986
	  & -- & -- & --
	  & -- & -- & -- \\
    PFT$_{\text{Whitening}}$
	  & 0.8625 & 0.8636 & 0.7416
	  & -- & -- & --
	  & -- & -- & -- \\
    PFT$_{\text{NormalizingFlow}}$
	  & 0.5620 & 0.6672 & 0.4465
	  & -- & -- & --
	  & -- & -- & -- \\
    REZE
	  & 0.9856 & 0.9667 & 0.9770
	  & -- & -- & --
	  & -- & -- & -- \\
    \hline

    \multicolumn{10}{l}{\emph{Qwen3-Embedding-0.6B}} \\
    FT
	  & 0.9045 & 0.8790 & 0.9124
	  & -- & -- & --
	  & -- & -- & -- \\
    PFT
	  & 0.9742 & 0.9900 & 0.9897
	  & -- & -- & --
	  & -- & -- & -- \\
    PFT$_{\text{Whitening}}$
	  & 0.7639 & 0.8131 & 0.8265
	  & -- & -- & --
	  & -- & -- & -- \\
    PFT$_{\text{NormalizingFlow}}$
	  & 0.9987 & 0.9912 & 0.7752
	  & -- & -- & --
	  & -- & -- & -- \\
    REZE
	  & 0.9430 & 0.9251 & 0.9759
	  & -- & -- & --
	  & -- & -- & -- \\
    \hline
  \end{tabular}
  \caption{
  Main results (2) on ChemTEB across different training sample sizes (100/500/1000).
  Each entry reports the score over the selected tasks within the corresponding benchmark.
  }
  \label{tab:per_task_results_5}
\end{table*}

\end{document}